\documentclass[10pt,twocolumn,letterpaper]{article}

\usepackage{iccv}
\usepackage{times}
\usepackage{epsfig}
\usepackage{graphicx}
\usepackage{amsmath}
\usepackage{amssymb}
\usepackage{multirow}
\usepackage{subcaption}

\usepackage[pagebackref=true,breaklinks=true,letterpaper=true,colorlinks,bookmarks=false]{hyperref}

\iccvfinalcopy 

\ificcvfinal\fi

\begin{document}

\title{Self-supervised Image-text Pre-training With Mixed Data In Chest X-rays}

\author{Xiaosong Wang, Ziyue Xu, Leo Tam, Dong Yang, Daguang Xu \\
Nvidia Corporation, Bethesda, MD\\
{\tt\small \{xiaosongw,ziyuex,leot,dongy,daguangx\}@nvidia.com}
}

\maketitle
\ificcvfinal\thispagestyle{empty}\fi

\begin{abstract}
   Pre-trained models, \eg, from ImageNet, have proven to be effective in boosting the performance of many downstream applications. It is too demanding to acquire large-scale annotations to build such models for medical imaging. Meanwhile, there are numerous clinical data (in the form of images and text reports) stored in the hospital information systems. The paired image-text data from the same patient study could be utilized for the pre-training task in a weakly supervised manner. However, the integrity, accessibility, and amount of such raw data vary across different institutes, \eg, paired vs. unpaired (image-only or text-only). In this work, we introduce an image-text pre-training framework that can learn from these raw data with mixed data inputs, \ie, paired image-text data, a mixture of paired and unpaired data. The unpaired data can be sourced from one or multiple institutes (\eg, images from one institute coupled with texts from another).
   Specifically, we propose a transformer-based training framework for jointly learning the representation of both the image and text data. In addition to the existing masked language modeling, multi-scale masked vision modeling is introduced as a self-supervised training task for image patch regeneration. We not only demonstrate the feasibility of pre-training across mixed data inputs but also illustrate the benefits of adopting such pre-trained models in 3 chest X-ray applications, \ie, classification, retrieval, and image regeneration. Superior results are reported in comparison to prior art using MIMIC-CXR, NIH14-CXR, and OpenI-CXR datasets.
\end{abstract}

\section{Introduction}
The development of recent machine learning paradigms in computer vision largely leverage the power of data, especially relying on the availability of large-scale annotated vision datasets for the supervised training, \ie, ImageNet~\cite{deng2009imagenet} (12 million images), MS COCO~\cite{lin2014microsoft} (1.5 million image objects),  Places~\cite{zhou2014learning} (11 million images), Visual Genome~\cite{krishnavisualgenome} (1.5 million object relationship instances), and JFT-300M~\cite{sun2017revisiting} (300+ million images). However, it also becomes the obstacle of developing and applying the most advanced learning techniques in certain specific fields at a similar scale, \eg, medical imaging. It is not cost-effective and often impossible to acquire such a large amount (more than thousands) of annotations from medical professionals. On the other side, tons of clinical data (in the form of images and text notes/reports) are accumulated via daily routine and then slept in the healthcare facilities' databases. These data are often not in the form that can be directly utilized but could potentially be valuable for learning and further benefit the future diagnosis of patients.

\begin{figure}[t]
    \centering
	\includegraphics[width=\columnwidth]{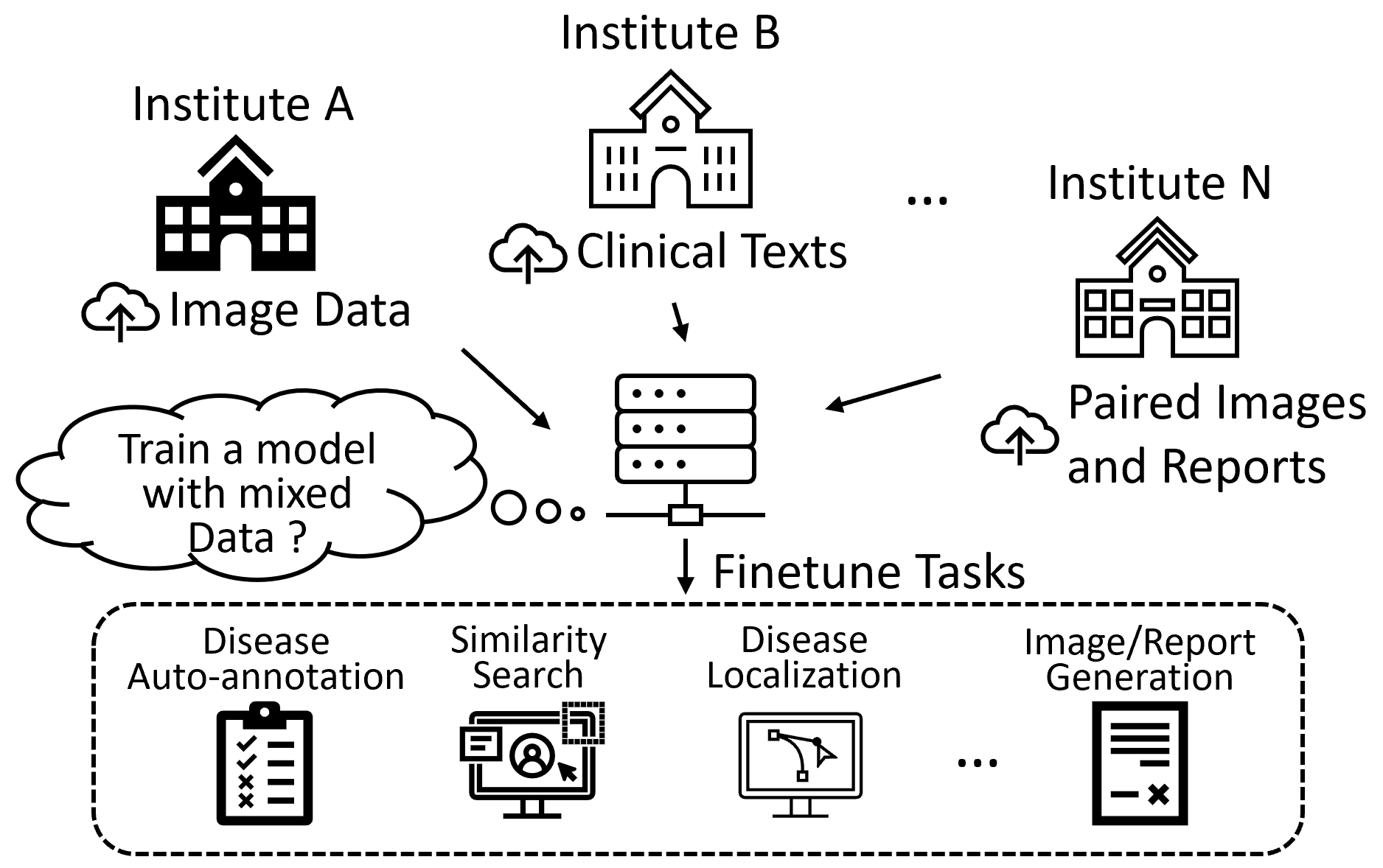}
	\caption{Overview of leveraging all accessible data.
	}
	\label{fig:motivation}
\end{figure}

Natural Language Processing (NLP) is another prosperous field in the rapid growth of modern deep learning techniques. In comparison to computer vision, NLP methods depend more on the representation learning in an unsupervised or self-supervised manner, \eg, word2vec~\cite{mikolov2013efficient}, Recurrent Neural Networks (RNN)~\cite{mikolov2010recurrent}, and most recent transformer models~\cite{vaswani2017attention,devlin2018bert,radford2019language}. The data themselves also serve as the supervision in a regeneration task. This way of leveraging the power of large-scale data via self-supervised learning without any annotations is particularly sound for the learning tasks in medical image analysis.

In addition, medical data usually contains more information that could be naturally utilized in the model training. The logical connection between the image and associated text (which usually comes with the image) from the same study and patient could be utilized as another level of supervision. However, we face the problem in practice that the integrity, accessibility, and amount of these data varies from institute to institute due to technical (out-dated equipment) and ethical (patient privacy protection) limits. Such data variability can pose a major challenge for learning from several different datasets via collaborative learning techniques like federated learning~\cite{konevcny2016federated,brisimi2018federated,yang2021federated}. Taking Chest X-rays for an example, we find several hospital-scale (in the scale of hundreds of thousands) datasets that are publicly accessible, \eg, NIH14-CXR~\cite{wang2017chestx}, CheXpert~\cite{irvin2019chexpert}, MIMIC-CXR~\cite{johnson2019mimic},  PadChest~\cite{bustos2020padchest}, and OpenI-CXR~\cite{demner2015preparing}. All of them provide the chest X-ray images (frontal and/or lateral views) with dataset-specific finding labels. In addition to image-label pairs, MIMIC-CXR, PadChest, and OpenI-CXR release the radiological reports associated with the images (PadChest's reports are in Spanish). The different ways of composing and releasing datasets make it hard for researchers to use all available data for the aimed analysis. 

Taking the initiative of `using all we have' (as shown in Fig.~\ref{fig:motivation}), we investigate how mixed data could be utilized for training a general and representative model. This work introduce an image-text pre-training framework that can handle the learning from these raw data with mix-up data inputs, \ie, paired image-text data, a mixture of paired and unpaired data. The unpaired data can be sourced from one or multiple institutes (\eg, images from one institute coupled with texts from another). A transformer-based unified training framework is designed for learning the representation of both the image and text data. In addition to the existing masked language modeling, multi-scale masked vision modeling is introduced as a self-supervised training task for image patch regeneration. Furthermore, the correlation between image and text among paired and unpaired data are utilized via our proposed cross-correlation modules. We believe it could be easily adopted and achieve better performance in various applications by lifting the demanding data requirement of training parameter-heavy models. Here, we not only demonstrate the feasibility of pre-training across mixed data inputs but also illustrate the benefits of adopting such pre-trained models in 3 chest X-ray applications, \ie, classification, retrieval, and image regeneration. Superior results are reported in comparison to prior arts using MIMIC-CXR, NIH14-CXR, and OpenI-CXR datasets.

In this work, our contributions are four-fold:
(1) We propose a novel transformer-based pre-training framework with various and mix-up data inputs for joint  representation learning of image and text in a self-supervised manner; 
(2) We propose a multi-scale masked vision model for the image data to complement the commonly used masked language model;
(3) We propose two approaches to model the correlation between image and text data that are either paired or unpaired;
(4) we apply the pre-trained models in chest X-rays and demonstrate the benefits of using the pre-trained model in fine-tuning tasks, especially when the annotated application dataset is small.

\section{Related Works}

ImageNet~\cite{deng2009imagenet} pre-trained Convolutional Neural Network (CNN) models have provided a general image representation and reinforced many downstream tasks, \eg, object detection~\cite{ren2015faster}, semantic/instance segmentation~\cite{he2017mask}, vision-language modeling~\cite{xu2015show}. There are a few studies~\cite{he2019rethinking,kolesnikov2019big} that demonstrate the effectiveness of using the pre-trained model on fine-tuning tasks with a relatively smaller dataset. On the NLP side, pre-training is indeed a fundamental technique for the modern NLP approaches, starting from word embeddings~\cite{mikolov2013efficient,pennington2014glove} to recent transformer-based language modeling, \ie, BERT\cite{devlin2018bert}, GPT~\cite{radford2018improving}, XLNet~\cite{yang2019xlnet}), which are all based on the pioneering work of ``Attention is all your need"~\cite{vaswani2017attention}.
Recently, pre-training models for vision-language tasks have been popular and largely leverage the transformer models designed for modeling sequence data. Typically, CNN extracted object features are appended to the word embedding sequence and processed with a shared transformer encoder. A variety of learning objectives are designed in addition to the masked word prediction task, \eg, masked region feature regression~\cite{chen2020uniter,wei2020multi}, text generation~\cite{sariyildiz2020learning,desai2020virtex,chen2021visualgpt,cho2021unifying}, and object classification~\cite{su2019vl,lin2020interbert}. The binding of image and text features are enforced via shared transformer (image and text in one sequence)~\cite{li2019visualbert,chen2020uniter,wei2020multi,su2019vl,lin2020interbert},  cross-attention~\cite{tan2019lxmert,lu2019vilbert}, and image-text mix-match~\cite{wei2020multi,lin2020interbert}. Zhang et al.~\cite{zhang2020contrastive} proposed a pre-training process with an image-text contrastive loss to align the paired image and text features (CNN features and BERT features) using chest X-ray datasets. From a technical perspective, the proposed mix-up pre-training framework is different from previous works in: (1) image patches regeneration (the image itself as supervision) instead of feature regression using a multi-scale image encoder and decoder; (2) image and text feature alignment via computing a correlation matrix using a mix-up of paired and unpaired data; (3) decoupled image and text encoder and decoder, which can facilitate the testing and fine-tuning task for either modality.


\section{Mix-up Image-text Pre-training}
\begin{figure*}[t]
    \centering
	\includegraphics[width=0.9\linewidth]{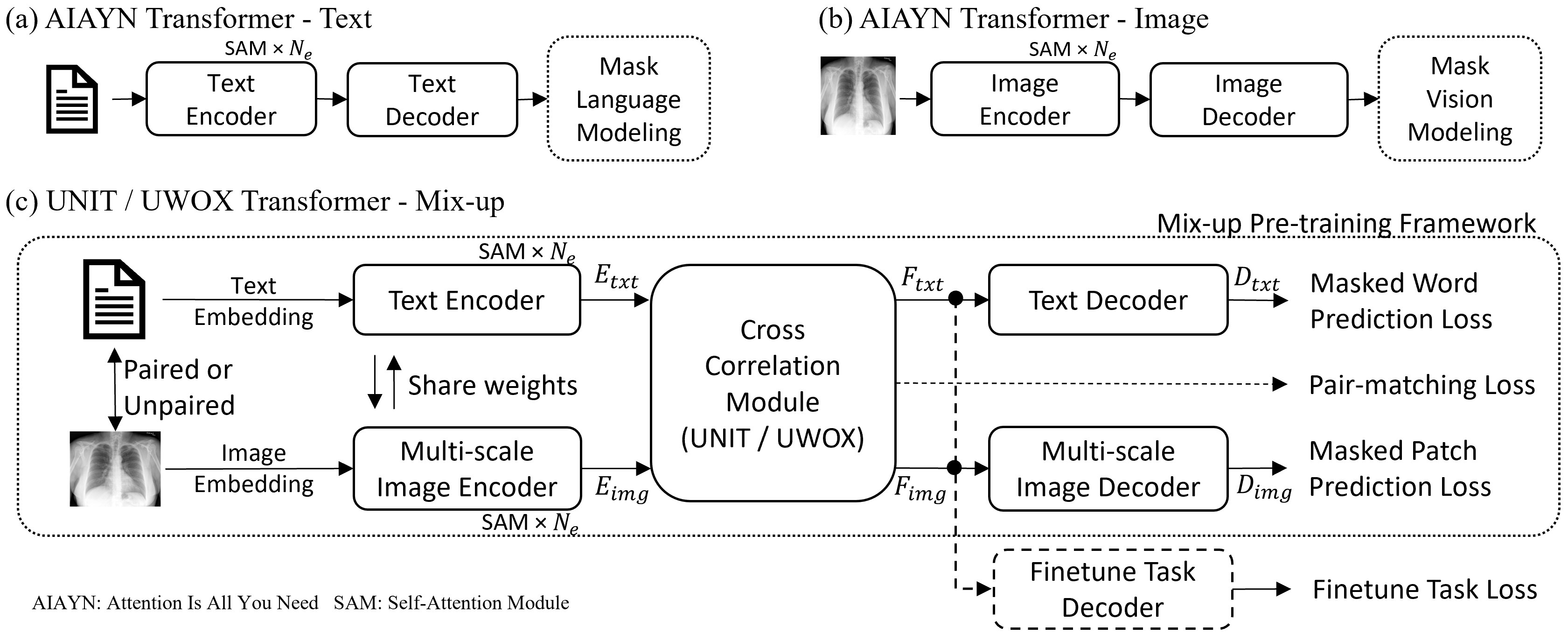}
	\caption{Overview of the proposed mix-up image-text pre-training framework.
	}
	\label{fig:framework}
\end{figure*}

The self-supervision feature of the NLP approaches is desirable for medical image data so as to lift the burden of manual annotations. A straightforward solution is to build transformer models for the medical images so that it can learn the representation of images in a similar fashion to the textual ones via the existing transformers. Nevertheless, images are not presented in the form of sequential data in comparison to textual data. Recent attempts~\cite{chen2020generative,dosovitskiy2020image} on image transformers use positional encoding to model the spatial information, which has space for further improvement.

In addition to separately modeling image and text, the correlation between the paired image and text can also be exploited for better representation learning. Existing methods mostly model such correlation as an extra step in a decoupled manner, while in this work, we postulate that in order to facilitate the interaction between image and text features, it will be more effective to learn and represent them in similar formats using a unified framework. Joint learning of image and text features will help connect the related phrases and image patches (often containing disease or diagnosis information) and therefore distinguish them from the rest of the redundant information that is shared by all data entries, \eg, image background and extraneous words. 

In this section, we will first discuss the baseline framework to model the image and text jointly with a unified transformer. Then, we illustrate two approaches, \ie, UNIT (UNIfied Transformer) and UWOX(UNIT WithOut Cross fusion), to model the correlation between image and text data. In the end, the proposed multi-scale masked vision model will be introduced to enhance the image encoding and decoding process.

\subsection{Baseline Framework}
Similar to BERT~\cite{devlin2018bert} and vision transformer~\cite{chen2020generative,dosovitskiy2020image}, we adopt the overall architecture of the transformer model from \cite{vaswani2017attention}. Fig.~\ref{fig:framework} illustrates the overview of our proposed learning framework (in Fig.~\ref{fig:framework}(c)) together with two baselines, `Attention is all you need' (AIAYN) transformers model for text (in Fig.~\ref{fig:framework}(a)) and image (in Fig.~\ref{fig:framework}(b)) individually. These two baseline models share the same transformer architecture while taking different inputs, either text ($X^t$) or image ($X^i$). $X^t$ represents a sequence of V words $X^t=\{x^t_1,\cdots,x^t_v,\cdots, x^t_V\}$ from each text data entry. On the image side, $X^i=\{x^i_1,\cdots,x^i_u,\cdots, x^i_U\}$ represents a sequence of flattened image patches $x^i_u$ with a length of $B\times B$. $B\times B$ also represents the image patch size before being flattened. Taking an image with a size of $256\times 256$ and $B=16$ as an example, the length of each $x^i_u$ is $256$ and there is totally a sequence of $U=256$ such flattened patches from a 2D image. Similar to the positional encoding $p^t_v$ for each $x^t_v$ (that indicates the index of words in each entry), a positional embedding will be generated for each image data feature $x^i_u$, which indicates the relative coordinates of top-left and bottom-right corners of each image patch, \ie, $[x_{start},y_{start},x_{end},y_{end}]$. The final embedding is defined as $\hat{X}^t=\{\hat{x}^t_v\}_V$ and $\hat{X}^i=\{\hat{x}^i_u\}_U$,
\begin{align}
    &\hat{x}^t_v=\mathrm{Norm}(W_x^t x^t_v + b^t_x)
    +\mathrm{Norm}(W^t_p p^t_v+b^t_p) \\
    &\hat{x}^i_u=\mathrm{Norm}(W_x^i x^i_u + b^i_x)
    +\mathrm{Norm}(W^i_p p^i_u+b^i_p)
    \label{eq:embedding}
\end{align}
where $\mathrm{Norm}$ is the layer normalization.

\begin{figure*}[t]
    \centering
	\includegraphics[width=0.95\textwidth]{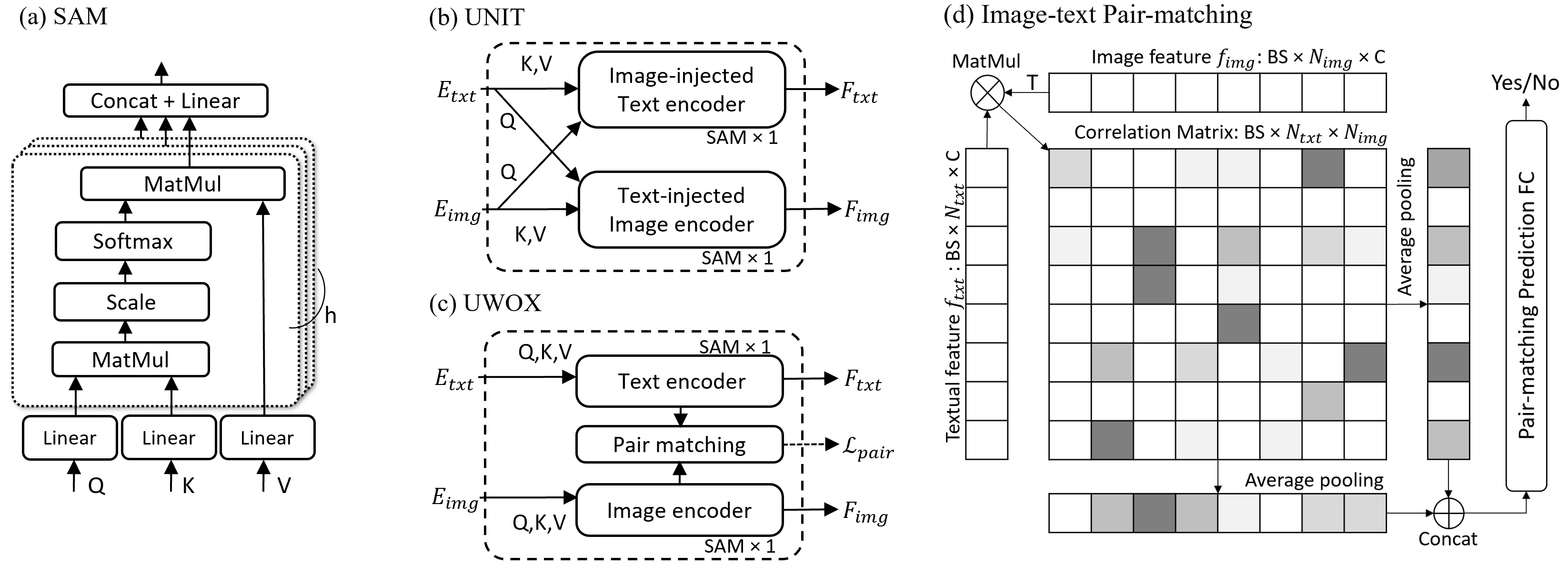}
	\caption{Components in the proposed framework: (a) SAM~\cite{vaswani2017attention}; (b) UNIT; (c) UWOX; (d) pair-matching module.
	}
	\label{fig:components}
\end{figure*}

As shown in Fig.~\ref{fig:framework}, each modality goes through an encoder-decoder based transformer framework with encoder for the representation learning and the decoder for the masked predictions of either image patches or textual words. The encoders are composed of a number ($N_e$) of multi-head Self-Attention Modules (SAM)~\cite{vaswani2017attention}. Fig.~\ref{fig:components}(a) shows the detailed structure of an encoding component $E=\mathrm{SAM}(Q,K,V)$, where $Q=K=V\in \{\hat{X}^t,\hat{X}^i\}$. For the baseline AIAYN transformers, we define a simple decoder as a 2-layer Multi-Layer Perceptron (MLP). During the training, a percentage of words or image patches will be masked out in the input (replaced with random words or image patches). They will be predicted in the end as a form of self-supervision. The loss $\mathcal{L}_{txt}$ for masked word prediction is defined as a multi-class cross-entropy loss, and $L_1$ Norm is employed for measuring the intensity differences as the $\mathcal{L}_{img}$ loss between predicted image patches and original ones. The total loss for the proposed mix-up pre-training framework is defined as $\mathcal{L}=\mathcal{L}_{txt}+\mathcal{L}_{img}+\mathcal{L}_{co}$, where $\mathcal{L}_{co}$ is the pair-matching loss that is optional and will be defined in the following section.

\subsection{Cross Correlation of Image and Text}
The proposed mix-up pre-training framework is designed as a combination of two baseline transformers for both image and text. Instead of simply taking the output from both models,  we align these two features by sharing the weights in the encoder modules $\mathrm{SAM}^{\times N_e}_e$ for image and text, which is proved to be a more effective and accurate way to model both modalities in our experiments.  We define the input as a tuple of image and text ($X^i$, $X^t$). $X^i$ and $X^t$ in the tuple could be either paired or unpaired. As we discussed above, we exploit the underlying correlation between paired image and text data. Thus, we experiment with two modules (UNIT and UWOX) to emphasize this connection for jointly learning image and text representations. The idea of UNIT (as shown in Fig.~\ref{fig:components}(b)) is similar to the cross-attention module presented in~\cite{tan2019lxmert}. It fuses the image and text features using a $\mathrm{SAM}$ module,
\begin{align}
    &F_{txt}=\mathrm{SAM}^{\times 1}_{unit}(E_{img},E_{txt},E_{txt}) \\
    &F_{img}=\mathrm{SAM}^{\times 1}_{unit}(E_{txt},E_{img},E_{img}).
    \label{eq:unit}
\end{align}
However, a transformer with the UNIT module is constraint by the requirement of the input containing both image and text for training and testing, which makes the pre-trained model less flexible for scenarios of single modality, either image or text alone. To remove this constraint, we propose to decouple the image-text fusion and strengthen the image-text correlation only during the training. In UWOX (as illustrated in Fig.~\ref{fig:components}(c)), $E_{txt}$ and $E_{img}$ are processed via a shared $\mathrm{SAM}_{uwox}^{\times 1}$ to produce $F_{txt}$ and $F_{img}$ without the fusion. Meanwhile, we compute the cross-correlation pair-matching loss during the training for either paired or unpaired image-text tuples.
\begin{align}
    &\mathrm{CoMat}= \mathrm{MatMul}(E_{txt}, E_{image}^{\mathrm{T}}) \\
    &\mathrm{Co}F_{img}=\mathrm{AvgPool(\mathrm{CoMat}, axis\mathrm{=}img)} \\
    &\mathrm{Co}F_{txt}=\mathrm{AvgPool(\mathrm{CoMat}, axis\mathrm{=}txt)} \\
    & \hat{\mathrm{I}}_{pair} = \mathrm{Sigmoid}(W_{co}(\mathrm{Co}F_{txt} \oplus \mathrm{Co}F_{img})+b_{co}) \\
    &\mathcal{L}_{co} = \mathrm{BCE}(\hat{\mathrm{I}}_{pair}, \mathrm{I}_{pair})
    \label{eq:uwox}
\end{align}
where $\mathrm{MatMul}$, $\mathrm{T}$, $\mathrm{AvgPool}$, $\mathrm{sigmoid}$ and $\oplus$ are the matrix multiplication operation, matrix transposing operator, average pooling operation for a 2D matrix on either directions, sigmoid function, and feature concatenation respectively. $\mathrm{BCE}$ stands for the Binary Cross Entropy loss and $\mathrm{I}_{pair}$ is the ground truth of whether the input ($X^i$, $X^t$) is paired or not ($1$ or $0$). This process is also detailed in Fig.\ref{fig:components}(d). When the data are all paired, $\mathcal{L}_{co}$ will be still be calculated but have less influence to the learning.  

\subsection{Multi-scale Image Encoding and Decoding}
\begin{figure*}[t]
    \centering
	\includegraphics[width=0.8\linewidth]{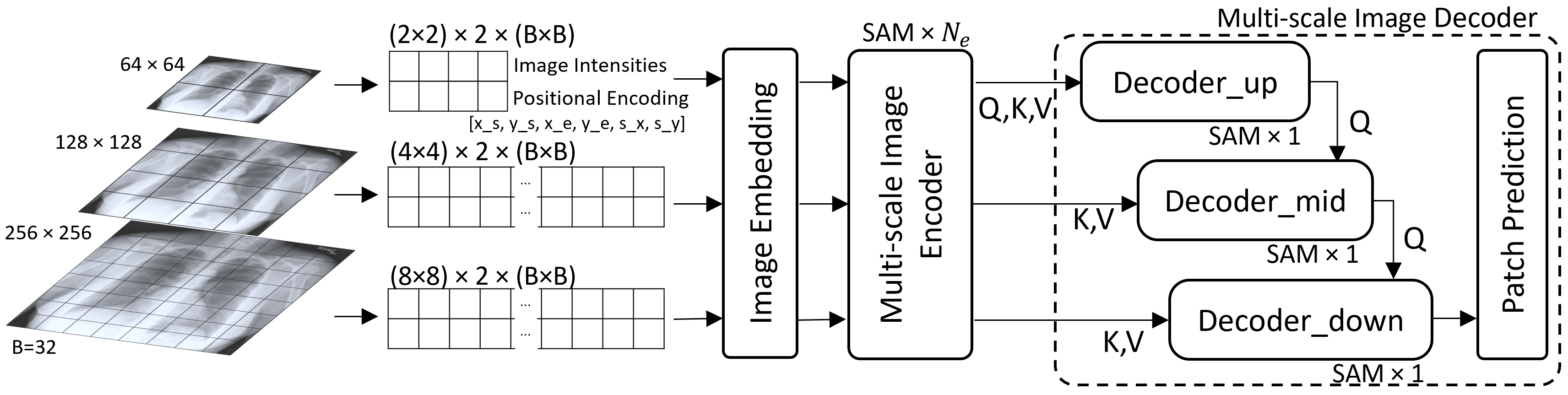}
	\caption{Detailed pipeline of multi-scale image encoding and decoding.
	}
	\label{fig:Multiscale_image}
\end{figure*}
The positional encoding used by existing works for image patch locations may not be sufficient to preserve the spatial information of images. This is because nearby image patches are often closely related, but direct patch predictions solely based on the features of each cell are not effective enough, which often lead to blocking artifacts (as shown in the result section later). It is one of the natural disadvantages of transformer-based methods for image processing compared to CNN-based ones. Here, we attempt to suppress this shortcoming within the general framework of transformer architectures. In Fig.~\ref{fig:Multiscale_image}, we illustrate a multi-scale image encoding and decoding process.   

For each image input $X^i$, we first build a 3-layer image pyramid with the original image as the bottom layer $X^{i,down}$, and reduce the image size by half in each dimension every time to construct the middle $X^{i,mid}$ and top layers $X^{i,up}$. For the positional encoding at each image scale, additional scale information will be included, \ie, $[x_{start},y_{start},x_{end},y_{end},x_{scale},y_{scale}]$, where $x_{scale}=y_{scale}\in\{0.25,0.5,1.0\}$. Similar to single scale image, we can embed the inputs into $\hat{X}^{i,up}$, $\hat{X}^{i,mid}$, and $\hat{X}^{i,down}$. Instead of encoding the single image $E_{img}=\mathrm{SAM}^{\times N_e}_e(\hat{X}^i,\hat{X}^i,\hat{X}^i)$, the framework will output
\begin{equation}
\begin{aligned}
    E_{img}^{S}=\mathrm{SAM}&^{\times N_e}_e(\hat{X}^{i,S},\hat{X}^{i,S},\hat{X}^{i,S}) \\
    &S \in \{up, mid, down\}
\end{aligned}
\label{eq:ms_encode}
\end{equation}
After computing $E_{img}^{S}, S\in\{up, mid, down\}$, we concatenate them together and further encode them via either UNIT or UWOX as $F_{img}^{S}$, $S\in\{up, mid, down\}$. Then, we output the decoded feature $D_{img}^{S}$, $S\in\{up, mid, down\}$ by propagating the upper layer features into the lower ones,
\begin{align}
    D_{img}^{up}=&\mathrm{SAM}^{\times 1}_d(F_{img}^{up},F_{img}^{up},F_{img}^{up}) \\
    D_{img}^{mid}=&\mathrm{SAM}^{\times 1}_d(D_{img}^{up},F_{img}^{mid},F_{img}^{mid}) \\
    D_{img}^{down}=&\mathrm{SAM}^{\times 1}_d(D_{img}^{mid},F_{img}^{down},F_{img}^{down})
    \label{eq:ms_decode}
\end{align}
Similarly, a 2-Layer MLP is integrated at the end to predict the masked image patches in $X^{i,down}$ using the decoder output $D_{img}^{down}$. 

In such a way, we can propagate the spatial information contained in upper scale patches into lower scale ones since each patch from the upper scale is closely related to at least four patches in the lower scale under the current setting. 

\begin{table*}[t]
\centering
\resizebox{\textwidth}{!}{
\begin{tabular}{|l|c|c|c|c|c|c||c|c|c|c|c|c|c|}
\multicolumn{14}{l}{\textbf{Baseline Scenarios}: Pre-train and fine-tune (left columns) or directly train (right ones) with Paired data (\textbf{P}: all training data in MIMIC)} \\
\hline
& \multicolumn{6}{c||}{Baseline Scenario 2} & \multicolumn{7}{c|}{Baseline Scenario 1}\\
\hline
OpenI-AUC   & AIAYN       & AIAYN      & \multicolumn{2}{c|}{UNIT}        & \multicolumn{2}{c||}{UWOX} &  \multicolumn{2}{c|}{UNIT-cls}  &  \multicolumn{2}{c|}{UWOX-cls} &      \multicolumn{2}{c|}{ResNet50}    &  TieNet         \\
\hline
disease                       & img\_only         & txt\_only            & img+txt            & txt+img            & img\_only            & txt\_only       & img+txt            & txt+img     & img\_only            & txt\_only    & img\_only            & img\&txt        & img\&txt        \\
\hline
Atelectasis                   & 0.776       & 0.961          & 0.944	&0.939	&0.773	&0.958         &0.927	&0.891  &0.688	&0.928 & 0.781          & 0.971          & 0.966          \\
Cardiomegaly                  & 0.890       & 0.952          & 0.942	&0.954	&0.903	&0.945         &0.896	&0.902 &0.747	&0.919 & 0.859          & 0.974          & 0.914          \\
Consolidation                 & 0.845       & 0.897          & 0.903	&0.917	&0.840	&0.953         &0.960	&0.961 &0.728	&0.950 & 0.829          & 0.876          & 0.920          \\
Edema                         & 0.885       & 0.994          & 0.994	&0.994	&0.923	&0.995         &0.977	&0.977 &0.821	&0.992 & 0.895          & 0.943          & 0.961          \\
E-cardio                      & 0.694       & 0.677          & 0.832	&0.808	&0.735	&0.681         &0.781	&0.801 &0.687	&0.762 & 0.795          & 0.538          & 0.630          \\
Fracture                      & 0.651       & 0.941          & 0.934	&0.937	&0.609	&0.962         &0.794	&0.826 &0.595	&0.840 & 0.513          & 0.778          & 0.931          \\
Lesion                        & 0.596       & 0.946          & 0.925	&0.920	&0.563	&0.925         &0.841	&0.847 &0.547	&0.900 & 0.585          & 0.799          & 0.951          \\
Opacity                       & 0.695       & 0.898          & 0.866	&0.879	&0.739	&0.898         &0.872	&0.881 &0.691	&0.899 & 0.742          & 0.917          & 0.919          \\
No-finding                    & 0.728       & 0.868          & 0.888	&0.897	&0.764	&0.881         &0.874	&0.880 &0.691	&0.866 & 0.754          & 0.889          & 0.882          \\
Effusion                      & 0.885       & 0.984          & 0.981	&0.983	&0.893	&0.979         &0.973	&0.973 &0.809	&0.971 & 0.912          & 0.959          & 0.956          \\
Pleural-other                 & 0.636       & 0.933          & 0.932	&0.914	&0.714	&0.904         &0.827	&0.845 &0.612	&0.803 & 0.648          & 0.735          & 0.902          \\
Pneumonia                     & 0.813       & 0.986          & 0.988	&0.989	&0.799	&0.988         &0.954	&0.962 &0.641	&0.985 & 0.781          & 0.897          & 0.941          \\
Pneumothorax                  & 0.658       & 0.936          & 0.968	&0.977	&0.798	&0.980         &0.959	&0.955 &0.646	&0.963 & 0.793          & 0.872          & 0.968          \\
Devices                       & 0.587       & 0.847          & 0.803	&0.849	&0.628	&0.865         &0.777	&0.764 &0.602	&0.792 & 0.628          & 0.783          & 0.873          \\
\hline
\textbf{AVG}              & 0.739     & 0.916     & 0.921       & \textbf{0.925}      & \textbf{0.763}       & \textbf{0.922}   &0.887	&0.890   &0.679	&0.898 & 0.751        & 0.852         & 0.908\\
\hline
\end{tabular}}
\caption{The classification results (averaged AUC) of our methods and previous arts with a common training scenario.}
\label{tab:cls-full}
\end{table*}

\section{Pre-training Scenarios and Applications}
In this section, we first discuss the datasets included in the experiments and then set up several common training scenarios so as to demonstrate our feasible solutions to the pre-training task with mixed data input.

\textbf{Datasets}: 
Two hospital-scale datasets (MIMIC-CXR and NIH14-CXR) are used for the training while we employed the OpenI-CXR dataset (with hand-labeled disease classes ground truth) as the universal testing set for all applications. \textbf{MIMIC-CXR}~\cite{johnson2019mimic} is a large-scale publicly available chest X-ray dataset. It contains 227,827 pairs of frontal-view X-ray images and free-text radiology reports. In total, 14 disease findings are annotated based on each report using NLP algorithms. Here, we adopted the training set of official patient splits. \textbf{NIH14-CXR}~\cite{wang2017chestx} is another publicly accessible chest X-ray dataset for thorax disease classification and localization, with a total of 112,120 frontal view images. It also contains 14 disease labels (different categories to MIMIC-CXR) mined from the associated reports while the reports are finally not released to the public. We also adopt the patient-level data splits published with the data for the training purpose in our experiments. \textbf{OpenI-CXR}~\cite{demner2015preparing} is a publicly shared chest X-ray dataset collected from multiple institutes by Indiana University. We adopt  3677 pairs of radiology reports and frontal-view images. The disease labels are hand-labeled, and we keep the same 14 finding categories as MIMIC-CXR for consistency. OpenI-CXR is employed here only for evaluation purpose.

\textbf{Baseline Scenarios 1}: There are a set of studies (paired images and reports) together with the associate annotations (\eg, image labels) available for the training in MIMIC-CXR. Pre-training is not performed. Models are directly trained from scratch for the applications.

\textbf{Baseline Scenarios 2}: Similar to baseline scenario 1, paired data are available with the annotation. However, a pre-trained model could be trained without using the annotation for obtaining a better image/text representation. Then, the learned features are utilized for the detail applications via a fast fine-tuning procedure. 

\textbf{Mix-up Scenarios 1}: Here, we simulated the situation that data are in different conditions using the data from a single institute (\ie, MIMIC-CXR). A fraction of them has paired image and text reports, while the others have images coupled with random reports. Only paired data has the associated annotation and will be further used for the fine-tuning of a specific application task. 

\textbf{Mix-up Scenarios 2}: It is similar to mix-up scenarios 1 but additional data are available from another institute (\ie, NIH14-CXR, which only has images). So the pre-training could be conducted using the mixed data from both institutes. Images from NIH14-CXR are coupled with random reports from MIMIC-CXR. This is the most realistic situation of collaborative learning across institutes for our proposed mix-up pre-training framework, which also serves as a snapshot of the ultimate scenario we mapped at the beginning. Again, only paired data that has the associated annotation will be used for fine-tuning the applications.

\textbf{Application Tasks}: In this work, we demonstrate three applications, 
\ie, disease classification, similarity search (patient study retrieval), and image regeneration. For each detailed fine-tuning task, a specific decoder needs to be trained in addition to the representation encoder. For the multi-label disease classification task, a global average pooling followed by a fully-connected layer is integrated to process $F_{img}$ and $F_{txt}$ separately and output the predictions for each class. So two sets of classification results will be presented for image and text. A BCE loss $\mathcal{L}_{cls}$ is employed for the fine-tuning. For the similarity search fine-tuning task, we append a fully connected layer to the 1-D pooled version of $F_{img}$ and $F_{txt}$ for generating a 64-bit hashing code and adopt the Cauchy hashing loss~\cite{cao2018deep} for the training. For each data entry in the OpenI-CXR, we retrieve all relevant cases (with the same disease labels) and rank the similarities based on the Hamming distance between two hash codes. 
Finally, we illustrate the performance of regenerating all the image patches for an unseen image. No additional training is required since image regeneration is indeed part of the pre-training framework. 

\section{Experiments}

\noindent\textbf{Evaluation Metrics}:
For the multi-label disease classification task, we compute the AUCs (Area Under the Curve) of Receiver Operating Characteristic (ROC) curves for each disease type. For the retrieval task, we report the retrieval precision by reporting $\mathrm{P@K}, \mathrm{K}\in\{1,5, 10, 50\}$. Only exact matching of the disease labels between two cases will count as a correct hit. For the image regeneration task, we use Mean Square Error (MSE), Peak Signal-to-Noise Ratio (PSNR), and Structural Similarity Index Measure (SSIM)~\cite{avcibas2002statistical} to measure the quality of regenerated images. 

\noindent\textbf{Implementation Details}: 
We use the same number of $\mathrm{SAM}$ as BERT model, \ie, $N_{e}=12$ in the proposed framework. The rates of masked patches and words are set to 15\% and fixed for all experiments. The size $C$ of hidden states in our transformers is also set to 768. A maximum length of 150 was used for all the reports. We use a common batch size $BS = 64$ and Adam optimizer (with a learning rate of 1e-4) for all the training with 2 NVIDIA Titan V 32G GPUs. For the pre-training, a BERT model (bert-base-uncased) is loaded to initialize the encoder weights.


\begin{table*}[t]
\centering
\resizebox{0.9\textwidth}{!}{
\begin{tabular}{|c|l|c|c|c|c|c|c|c|c|c|}
\multicolumn{11}{l}{\textbf{Mix-up Scenario 1}: Pre-train with mix-up of Paired (\textbf{P}: fraction of MIMIC) + unPaired(\textbf{uP}: rest of MIMIC) and finetune on \textbf{P}} \\
\hline
\textbf{OpenI}               & AVG AUC         & P1\%+uP    & P2\%+uP    & P3\%+uP    & P5\%+uP    & P10\%+uP   & P30\%+uP   & P50\%+uP   & P90\%+uP  & P100\% \\
\hline
\multirow{2}{*}{UNIT}        & img+txt    & 0.905       & 0.897       & 0.900       & 0.912       & 0.914     & 0.909     & 0.910     & 0.908   &-  \\
                             & txt+img    & 0.895       & 0.908       & 0.911       & 0.921       & 0.913     & 0.903     & 0.906     & 0.921   &-  \\
\hline
\multirow{2}{*}{UWOX}        & img\_only    & \underline{0.735}& \underline{0.753}& \underline{0.751}       & 0.747       & 0.749     & 0.753     & 0.760     & 0.768   &-  \\
                             & txt\_only    & 0.908       & 0.905       & 0.908       & 0.906       & 0.908     & 0.905     & 0.904     & 0.903 &-    \\
\hline 
\multicolumn{11}{l}{\textbf{Mix-up Scenario 2}: Pre-train with mix-up of Paired (\textbf{P}: fraction of MIMIC) + unPaired(\textbf{uP}: all NIH14) and then finetune on \textbf{P}}                  \\
\hline
OpenI                        & AVG AUC         & P1\%+uP    & P2\%+uP    & P3\%+uP    & P5\%+uP    & P10\%+uP   & P30\%+uP   & P50\%+uP   & P90\%+uP  &P100\%+uP \\
\hline
\multirow{2}{*}{UNIT}        & img+txt    & 0.893       & 0.911       & 0.897       & 0.915       & 0.915     & 0.909     & 0.922     & 0.920   & \textbf{0.926}  \\
                             & txt+img    & 0.885       & 0.906       & 0.905       & 0.922       & 0.918     & 0.908     & 0.909     & 0.923   & 0.921  \\
\hline
\multirow{2}{*}{UWOX}        & img\_only    & \underline{0.620}       & \underline{0.649}       & \underline{0.670}       & 0.696       & 0.708     & 0.734     & 0.739     & 0.761  & \textbf{0.765}   \\
                             & txt\_only    & 0.904       & 0.895       & 0.910       & 0.913       & 0.918     & 0.889     & 0.903     & 0.891    & 0.911 \\
\hline 
\multicolumn{11}{l}{\textbf{Baseline Scenario 1}: Directly train with Paired (\textbf{P}: fraction of MIMIC)}                                                                              \\
\hline
OpenI                        & AVG AUC         & P1\%       & P2\%       & P3\%       & P5\%       & P10\%      & P30\%      & P50\%      & P90\%     &P100\% \\
\hline
\multirow{2}{*}{ResNet50~\cite{wang2017chestx}}    & img\_only    & \underline{0.596}       & \underline{0.603}       & \underline{0.643}       & 0.661       & 0.664     & 0.720     & 0.724     & 0.751   & 0.751  \\
                             & img\&txt      & 0.705       & 0.743       & 0.752       & 0.770       & 0.787     & 0.800     & 0.808     & 0.826    & 0.852 \\
\hline
TieNet~\cite{wang2018tienet} & img\&txt      & 0.760       & 0.779       & 0.836       & 0.856       & 0.885     & 0.900     & 0.904     & 0.895    & 0.908 \\
\hline
\multicolumn{10}{l}{\textbf{Baseline Scenario 2}: Pre-train and finetune with Paired (\textbf{P}: fraction of MIMIC)}                                                                          \\
\hline
OpenI                        & AVG AUC         & P1\%       & P2\%       & P3\%       & P5\%       & P10\%      & P30\%      & P50\%      & P90\%    & P100\%  \\
\hline
\multirow{2}{*}{UNIT}        & img+txt    & 0.518       & 0.855       & 0.909       & 0.916       & 0.911     & 0.914     & 0.915     & 0.917   & 0.921  \\
                             & txt+img    & 0.580       & 0.888       & 0.911       & 0.913       & 0.914     & 0.919     & 0.913     & 0.919   & 0.925  \\
\hline
\multirow{2}{*}{UWOX}        & img\_only    & \underline{0.507}    & \underline{0.588}    & \underline{0.657}    & 0.676 & 0.701     & 0.724     & 0.742     & 0.756    &0.763 \\
                             & txt\_only    & 0.502       & 0.883       & 0.901       & 0.891       & 0.919     & 0.921     & 0.903     & 0.907     & 0.922\\
\hline

\end{tabular}}
\caption{It illustrates the classification results (AVG AUC) of our methods and previous arts with various settings of training scenarios, using the hand-labeled OpenI dataset. \underline{Underlined} numbers highlight the improvements across scenarios.}
\label{tab:mixup-cls}
\end{table*}

\subsection{Results and Discussion}
\noindent\textbf{Classification Results}: 
First, we evaluate the classification performance of our proposed framework in two baseline scenarios, in which the general benefit of performing pre-training could be illustrated. AIAYN (image and text) models and two variation of our proposed methods (UNIT and UWOX) are pre-trained and then fine-tuned (with $\mathcal{L}_{cls}$ only) using all the data in the MIMIC-CXR training set. Methods for comparison are trained using the same amount of annotated data. UNIT-cls and UWOX-cls show the results without pre-training. ResNet50 stands for a CNN based disease classifier~\cite{wang2017chestx}. For the classification using both image and text report (img\&txt), we directly extract the textual feature via a pre-trained Biobert~\cite{lee2020biobert} model and concatenate it together with the output of pool5 in ResNet-50 for the final classification. TieNet~\cite{wang2018tienet} represents the state-of-the-art method in the field, which is also based on ResNet-50 and a LSTM based text encoder. Table~\ref{tab:cls-full} shows the AUCs for each disease class and Average (AVG) AUCs for all models. `img+txt' represents the results by feeding the text embedded image feature $F_{img}$ to the classifier. `txt+img' is computed with $F_{txt}$. For this task, UNIT achieve the highest AUC using image and text fused features. The classification power is mainly from the text as shown, while adding image further promotes the performance. Our UWOX outperforms other images-only classifiers. Also, it performs better than the vision transformer based model AIAYN-image (same setting as \cite{dosovitskiy2020image} with patch prediction on a single scale). 


Next, we test our proposed mix-up pre-training work in two scenarios with a mixed set of training data, \ie, Mix-up Scenario 1 and 2. The size of Paired data (P, also represents the set of data with annotations) are varied with a list of percentages. 1\% of MIMIC-CXR training set has about 2000 pairs of image and text data. As shown in Table~\ref{tab:mixup-cls}, we sample more frequently at the smaller percentages (less paired and annotated data) since smaller percentages of annotated data have larger influence on the performance. 
We carefully ensure all the compared methods sharing the same paired and annotated data, while the methods in mix-up scenarios also learn from extra unpaired data 
during the pre-training stage. Our methods (UNIT and UWOX that see extra data during the pre-training) shows significantly higher AVG AUCs than the ones do not learn from the extra data, \ie Mix-up Scenarios 1\&2 vs. Baseline Scenario 2. The gap is larger when less annotated data are available (small percentages of Paired data) . It indicates that a model pre-trained with a large amount of (either paired or unpaired) data could help applications with small annotated datasets. We also notice that UWOX performs worse in Mix-up Scenario 2 than in Mix-up Scenario 1 when there is a much smaller amount of paired and annotated data compared to unpaired ones. This may be due to the domain gap between the image data from two institutes. However, we also note that the model pre-trained with entire MIMIC-CXR and NIH14-CXR achieves the overall best result (\ie, P100\%+uP in Mix-up Scenario 2).
\begin{figure*}[t]
    \centering
    
	\includegraphics[width=0.8
	\linewidth]{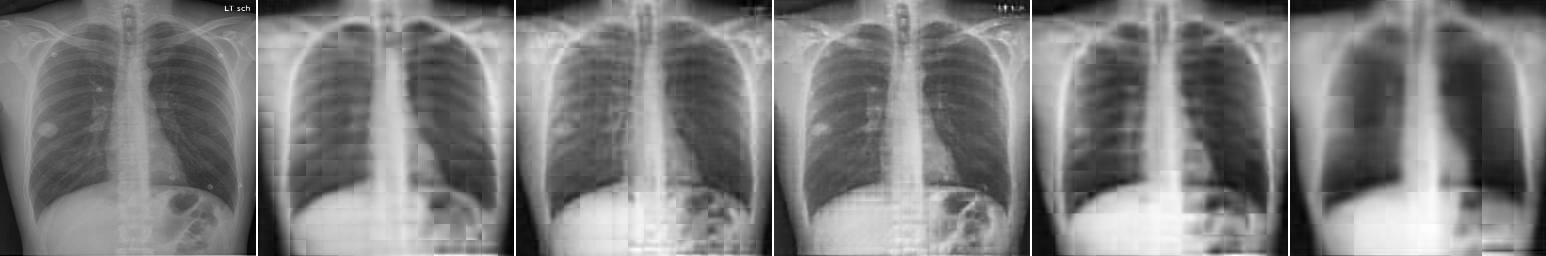}\\
	\includegraphics[width=0.8
	\linewidth]{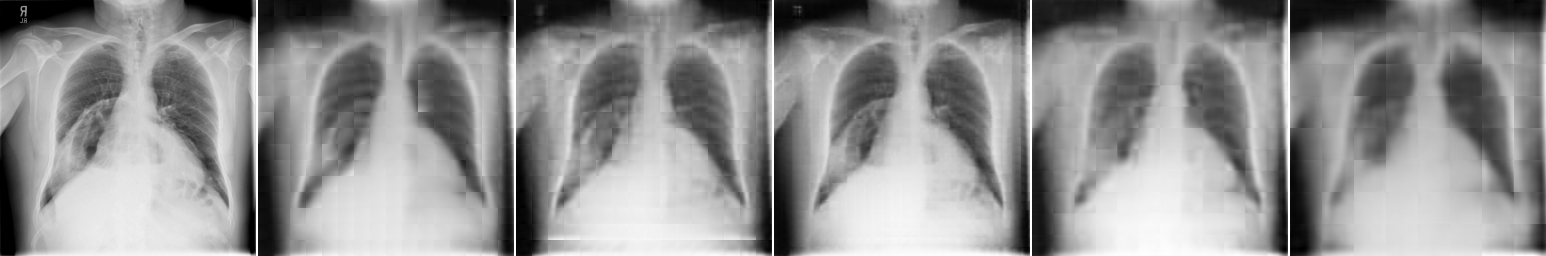}\\
	\includegraphics[width=0.8
	\linewidth]{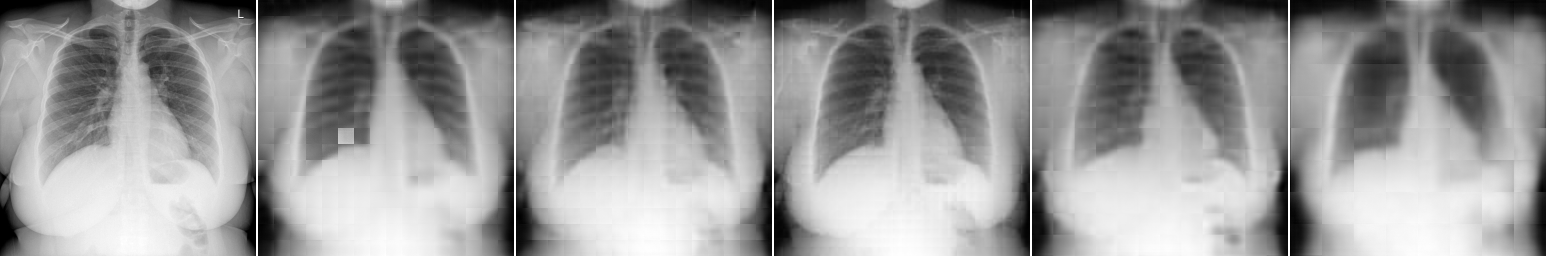}\\
	\includegraphics[width=0.8
	\linewidth]{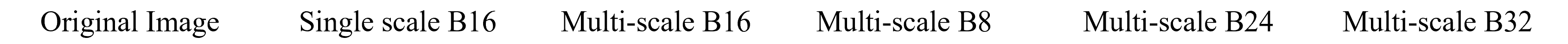}
	\caption{Image Regeneration results by UWOX with a variety of image block sizes. }
	\label{fig:image_recon}
\end{figure*}

\begin{table}[t]
\centering
\resizebox{0.73\columnwidth}{!}{
\begin{tabular}{|l|r|r|r|r|}
\hline
OpenI      & P@1 & P@5 & P@10 & P@50 \\
\hline
\multicolumn{5}{c}{Image-query based similarity search } \\
\hline
ResNet50~\cite{he2016deep} & 0.673	&0.546	&0.488	&0.503\\
UWOX       & 0.636	&0.533	&0.583	&0.507\\
UWOX-mixup & \textbf{0.687}	&0.657	&0.621	&0.570\\
\hline
\multicolumn{5}{c}{Text-query based similarity search}  \\
\hline
LSTM~\cite{wang2018tienet}       & 0.704	&0.566	&0.568	&0.487\\
Biobert~\cite{lee2020biobert}    & 0.698	&0.588	&0.534	&0.545 \\
UWOX       & 0.745	&0.738	&0.602	&0.653\\
UWOX-mixup & \textbf{0.763}	&0.733	&0.603	&0.657\\
\hline
\multicolumn{5}{c}{Image+Text-query based similarity search }    \\
\hline
TieNet~\cite{wang2018tienet}     &  0.702	&0.663	&0.547	&0.510 \\
UWOX       & 0.747	&0.746	&0.734	&0.670 \\
UWOX-mixup & 0.737	&0.731	&0.732	&0.640  \\
UNIT       & \textbf{0.782}	&0.590	&0.621	&0.637 \\
\hline
\end{tabular}}
\caption{Retrieval results with 3 query types.}
\label{tab:simseach}
\end{table}

\noindent\textbf{Similarity Search Results}: We evaluate the retrieval precision of our methods and previous works via three different query types, \ie, image-query, text-query, and image+text-query. Methods with textual features perform stronger than image-based ones in general. Here, UWOX is trained using all training data in MIMIC-CXR while UWOX-mixup additionally sees all images in NIH14-CXR. Our proposed methods demonstrate better accuracies in all three tasks.

\noindent\textbf{Image Patch Regeneration}:
Here, we vary the block size $B$ for image encoding and decoding to see how it will affect the image regeneration task. As shown in Fig.~\ref{fig:image_recon} and Table~\ref{tab:ablation}(b), block size $B=8$ renders the best image regeneration quality compared to other variants. Block size $B=16$ demonstrates higher classification accuracy with a moderate requirement of the computing resources (larger \# patches per image will require larger GPU memories). As shown in Fig.~\ref{fig:image_recon}, our proposed multi-scale image encoder and decoder can provide better regeneration results than a single-scale model. It generally suffers less from the blocking artifacts and is able to preserve more details, especially for pathological regions, as shown on the top two examples in Fig.~\ref{fig:image_recon}.

\begin{table}[t]


\begin{subtable}{0.48\textwidth}
\centering
\resizebox{\columnwidth}{!}{
\begin{tabular}{|l|l|c|c|c|}
\multicolumn{5}{l}{(a) UWOX with/without Pair-matching (Mix-up Scenario 1) } \\
\hline
OpenI - UWOX                                                                                              & AUC & P1\%+uP & P2\%+uP & P3\%+uP \\
\hline
\multirow{2}{*}{\begin{tabular}[c]{@{}l@{}}Train with image-text \\ pair-matching loss\end{tabular}} & img      & 0.735    & 0.755    & 0.751    \\
                                                                                                   & txt      & 0.908    & 0.905    & 0.908    \\
\hline
\multirow{2}{*}{\begin{tabular}[c]{@{}l@{}}Train without image-text \\ pair- matching loss\end{tabular}}                                                        & img  & 0.727    & 0.732    & 0.730    \\
                                                                                                   & txt  & 0.880    & 0.907    & 0.906    \\
\hline
\end{tabular}}
\label{tab:ablation_loss}
\end{subtable}%
\bigskip

\begin{subtable}{0.48\textwidth}
\centering
\resizebox{\columnwidth}{!}{
\begin{tabular}{|l|c|c|c|c|c|}
\multicolumn{6}{c}{(b) Comparison of block sizes in multiscale image encoding} \\
\hline
openI - UWOX      & ss\_B16~\cite{dosovitskiy2020image} & ms\_B16 & ms\_B8 & ms\_B24 & ms\_B32 \\
\hline
\multicolumn{6}{c}{Image patch regression (Baseline Scenario 2 P100\%)} \\
\hline
AVG MSE & 479.5 & 406.2 & 287.5 & 528.0 & 624.2 \\
AVG PSNR & 353.8 & 354.6 & 356.5 & 353.5 & 352.4 \\
AVG SSIM & 0.768 & 0.793 & 0.872 & 0.780 & 0.720 \\
\hline
\# patches/image & 256 & 336 & 1029 & 189 &84 \\
\hline
\multicolumn{6}{c}{Image Classification (Baseline Scenario 2 P100\%)}\\
\hline
AVG AUC-img & 0.739 & 0.763 & 0.761 & 0.744 & 0.703  \\
AVG AUC-txt  & 0.902 & 0.922 & 0.905 & 0.883 & 0.893 \\
\hline
\end{tabular}}
\label{tab:ablation_blocksize}
\end{subtable}%
\caption{Ablation studies on (a) pair-matching loss; (b) image block sizes. ss: single-scale; ms: multi-scale}
\label{tab:ablation}
\end{table}


\noindent\textbf{Effectiveness of Pair-matching Loss}:
Pair-matching loss is an essential component in our proposed UWOX transformer model, which learns the extra information from the underlying correspondence between image and text via the paired and unpaired comparison. In Table~\ref{tab:ablation}(a), we illustrate the performance boost between models trained with and without the pair-matching loss. The improvements are more obvious for the image-only case, considering that the results for the text features are already quite high. 

\noindent\textbf{Potentials of Transformers in Vision Tasks}:
In this work, we try to adopt only the transformer-based models/modules as a comparison to current CNN-based image models. Our proposed transformer for image modeling can achieve equivalent or better performance (with sufficient training data) to CNN-based ones in all three applications, while most of them do not rely much on the spatial information in images. Dosovitskiy et al.~\cite{dosovitskiy2020image} also report superior results of a transformer-based classifier when enough data are feed. The proposed multi-scale image encoder and decoder modules are proven to be effective in preserving the spatial information but also has the potential to be enhanced and extended for other applications, \eg, image pyramid with overlapped image blocks and image segmentation tasks.

\section{Conclusion}
In this work, we demonstrate how a mix-up set of image-text data could be utilized together for learning a more generalized pre-training model using a transformer-based architecture. We believe it is a feasible and valuable solution to learn and make use of the tremendous amounts of data that are already in our hospital databases.

{\small
\bibliographystyle{ieee_fullname}
\bibliography{egbib}
}

\end{document}